\documentclass[a4paper, 10pt, conference]{ieeeconf}      % Use this line for a4
\usepackage{FG2020}
\usepackage{xspace}
\usepackage{graphicx}
\usepackage{amsmath,amssymb} % define this before the line numbering.
\usepackage{color}
\usepackage{pifont}
\usepackage{booktabs}
\usepackage{lipsum}
\usepackage{xcolor}
\usepackage{multirow}
\usepackage{wrapfig}
\usepackage{tikz}
\usepackage{setspace}
\usepackage{pgfplots}
\pgfplotsset{compat=newest}
\usepackage{balance}
\usepackage{mathtools}
\usepackage{algorithm}
\usepackage{algorithmic}

\FGfinalcopy % *** Uncomment this line for the final submission

\IEEEoverridecommandlockouts                              % This command is only
\title{\LARGE \bf
Clustering based Contrastive Learning \\for Improving Face Representations
}

\author{\parbox{16cm}{\centering
    {\large Vivek Sharma$^{1,2}$, Makarand Tapaswi$^3$, M. Saquib Sarfraz$^{1,4}$ and Rainer Stiefelhagen$^1$}\\
    {\normalsize $^1$Karlsruhe Institute of Technology, $^2$Massachusetts Institute of Technology, $^3$Inria Paris, and $^4$Daimler TSS }\\
}}

\begin{document}

\makeatletter
\DeclareRobustCommand\onedot{\futurelet\@let@token\@onedot}
\def\@onedot{\ifx\@let@token.\else.\null\fi\xspace}

\def\eg{\emph{e.g}\onedot} \def\Eg{\emph{E.g}\onedot}
\def\ie{\emph{i.e}\onedot} \def\Ie{\emph{I.e}\onedot}
\def\cf{\emph{c.f}\onedot} \def\Cf{\emph{C.f}\onedot}
\def\etc{\emph{etc}\onedot} \def\vs{\emph{vs}\onedot}
\def\wrt{w.r.t\onedot} \def\dof{d.o.f\onedot}
\def\etal{\emph{et al}\onedot}
\makeatother
% ---------------------------------------------------------------
% ---------------------------------------------------------------

\def\bx{\mathbf{x}}
\def\bt{\mathbf{t}}
\def\bv{\mathbf{v}}

\ifFGfinal
\thispagestyle{empty}
\pagestyle{empty}
\else
\author{Anonymous FG2020 submission\\ Paper ID \FGPaperID \\}
\pagestyle{plain}
\fi
\maketitle

%%%%%%%%%%%%%%%%%%%%%%%%%%%%%%%%%%%%%%%%%%%%%%%%%%%%%%%%%%%%%%%%%%%%%%%%%%%%%%%%
\begin{abstract}
% \saquib{Really good job Makarand, admie your writing skills :), just a little note,I think 'learning to cluster' in the title is a bit misleading, how about something alternative like above, and we probably should term the CCL as something else as it sounds too much similar to the previous SSIAM or TSIAM, May be CCL for selfsupvised Contrastive Transfer??}
A good clustering algorithm can discover natural groupings in data.
These groupings, if used wisely, provide a form of weak supervision for learning representations.
In this work, we present Clustering-based Contrastive Learning (\emph{CCL}), a new clustering-based representation learning approach that uses labels obtained from clustering along with video constraints to learn discriminative face features.
We demonstrate our method on the challenging task of learning representations for video face clustering.
Through several ablation studies, we analyze the impact of creating pair-wise positive and negative labels from different sources.
Experiments on three challenging video face clustering datasets: BBT-0101, BF-0502, and ACCIO show that CCL achieves a new state-of-the-art on all datasets.
\end{abstract}

\section{Introduction}
\label{sec:intro}

% We address the problem of effectively learning a feature representation.
Learning strong and discriminative representations is important for diverse applications such as face analysis, medical imaging, and several other computer vision and natural language processing (NLP) tasks.
While considerable progress has been made using deep neural networks, learning a representation often requires a large-scale dataset with manually curated ground-truth labels.
To harness the power of deep networks on smaller datasets and tasks, pre-trained models (\eg~ResNet-101~\cite{he2016deep} trained on ImageNet, VGG-Face~\cite{vggface} trained on a large number of face images) are often used as feature extractors or fine-tuned for the new task.

% \makarand{why is the next sentence here? how does it concern this paper}
% Beyond the difficulty of creating large clean datasets, a lot of effort is invested in tackling other challenges such as class imbalance, model hyper-parameters, and the network architecture itself.
% There is no oracle method that works on all datasets.
% \makarand{reading the first para makes me feel like this paper will provide a new way to learn end-to-end without gt labels}

We introduce a clustering-based learning method \textit{CCL} to obtain a strong face representation on top of features extracted from a deep CNN.
An ideal representation has small distances between samples from the same class, and large inter-class distances in feature space.
% The learning of such representations can be categorized into different types based on the source of the ground-truth labels.
% In the \textit{fully-supervised} case ground-truth labels are available (often obtained through crowd-sourcing).
Different from a fully-supervised setting where ground-truth labels are often provided by humans, we view CCL as a \textit{self-supervised} learning paradigm where ground-truth labels are obtained automatically based on natural groupings of the dataset.
% In the \textit{fully-supervised} learning paradigm, the models can learn accurate margins and better separability between the target classes using cross-entropy loss.
% This reliance on labeling the ground-truth data is expensive as it requires manual labeling.

Self-supervised learning methods are receiving increasing attention as collecting large-scale datasets is extremely expensive.
In NLP, word and sentence representations (\eg~word2vec~\cite{word2vec}, BERT~\cite{bert}) are often learned by modeling the structure of the sentence.
In vision, there are efforts to learn image representations by leveraging spatial context~\cite{doersch2015}, ordering frames in a short video clip~\cite{fernando2017self,ref2}, or tracking image regions~\cite{ref3}.
There are also efforts to learn multimodal video representations by ordering video clips~\cite{sharma_hvu}.

Among recent works in the \textit{self-supervised} learning paradigm, \cite{caron2018deep,guo2017improved,jiang2016variational,xie2016unsupervised,yang2016joint} propose to learn image representations (often with an auto-encoder) using a clustering algorithm as objective, or as a means of providing pseudo-labels for training a deep model.
One challenge with the above methods is that the models need to know the number of clusters (usually the number of categories) a priori.
Our paper is related to the above as we utilize clustering algorithms to discover the structure of the data, however, we do not assume knowledge of number of clusters.
In fact, our method is based on working with a clustering setup that yields a large number of clusters with high purity and few samples per cluster.
% others where a priori knowledge of data is required. %, and we use the predicted cluster labels for training a neural network with an overall goal to improve video face clustering.

% propose to use predicted cluster labels or weak labels obtained from a clustering algorithm that has the potential of discovering groupings in the data with high purity.
Our main contribution, CCL, is a method for refining feature representations obtained using deep CNNs by discovering dataset clusters with high purity and typically few samples per cluster, and using these cluster labels as potentially noisy supervision.
In particular, we obtain positive (same class) and negative (different class) pairs of samples using the cluster labels and train a Siamese network.
We adopt the recently proposed FINCH clustering algorithm~\cite{finch} as a backbone since it provides hierarchical partitions with high purity, does not need any hyper-parameters, and has a low computational overhead while being extremely fast.
% Specifically, using these cluster labels is one of our contributions. In this paper, we use the predicted cluster labels for training a neural network with an overall goal to improve video face clustering.

Towards our task of clustering faces in videos, we demonstrate how cluster labels can be integrated with video constraints to obtain positive (within and across neighboring clusters) and negative pairs (co-occurring faces and samples from farthest clusters).
Both the cluster labels and video constraints are used to learn an effective face representation.

% (i) positive pairs from faces within a cluster;
% (ii) 
% The second contribution of the paper is how to complementarily integrate these weak cluster labels with the face detection (or \textit{video constraints}) such as (1) no two faces in the same frame could be assigned to the same cluster; (2) for frames with singleton (no co-occurring) face detection, we incorporate negative cluster pairs by randomly sample farthest cluster; (3) all faces in the cluster are considered positive pairs.  Our contribution is a method that jointly utilizes weak cluster labels and video constraints for learning an effective face feature representation. Namely, we call our method, \textit{USiam}.

Our proposed approach is evaluated on three challenging benchmark video face clustering datasets: Big Bang Theory~(BBT), Buffy the Vampire Slayer~(BF) and Harry Potter~(ACCIO).
We empirically observe that deep features can be further refined by using weak cluster labels and video constraints leading to state-of-the-art performance.
Importantly, we also discuss why the cluster labels may be more effective than using labels from alternative groupings such as tracking (\eg~TSiam~\cite{ssiam}).

The remainder of this paper is structured as follows:
Section~\ref{sec:relwork} provides an overview of related work.
In Section~\ref{sec:model}, we propose our method for clustering-based representation learning, CCL.
Extensive experiments, an ablation study, and comparison to the state-of-the-art are presented in Section~\ref{sec:eval}.
Finally, we conclude our paper in Section~\ref{sec:conclusion}.

\section{Related Work} \label{sec:relwork}

This section discusses face clustering and identification in videos.
We primarily present methods for face representation learning with/without CNNs, followed by a short overview of FINCH-clustering algorithm and learning from clustering.

\vspace{2mm}
\noindent\textbf{Video face clustering}
methods often follow 2 steps:
obtain pairwise constraints typically by analyzing tracks, followed by representation/metric learning approaches and clustering.
We present models through a historical perspective, with and without a CNN.

\vspace{2mm}
\noindent\textbf{Face representation learning without CNNs.}
The most common source of constraints is the \textbf{temporal information} contained in face tracks.
Faces within the same track are assigned a positive label (same character), while faces from temporally overlapping tracks are assigned a negative label (different characters).
This approach has been exploited by learning a metric to obtain cast-specific distances~\cite{cinbis2011unsupervised} (ULDML);
iteratively creating a generative clustering model and associating short sequences using a hidden Markov Random Field (HMRF)~\cite{wu2013simultaneous,wu2013constrained};
or performing clustering with a weighted block-sparse low-rank representation (WBSLRR)~\cite{xiao2014weighted}. 
% A matrix-factorization technique trying to reconstruct de-noised representations is presented in~\cite{r2}.
Additional constraints from \textbf{video editing cues} such as shots, threads, and scene boundaries are used in an unsupervised way to merge tracks~\cite{tc} by learning track and cluster representations with SIFT Fisher vectors~\cite{vf2}.

\vspace{2mm}
\noindent\textbf{Face representation learning with CNNs.}
With the popularity of Convolutional Neural Networks (CNNs), there is a growing emphasis on improving face track representations using CNNs.
An improved triplet loss that pushes the positive and negative samples apart in addition to the anchor relations is used by~\cite{imptriplet} to fine-tune a CNN.
% that in addition to the traditional triplet loss also push the positive and negative samples apart.
Zhang~\etal~\cite{jfac}~(JFAC) use a Markov Random Field to discover dynamic constraints iteratively to learn better representations during the clustering process.
Inverse reinforcement learning on a ground-truth data is also used to learn a reward function that decides whether to merge a given pair of face features~\cite{merge-or-not}.
In a slightly different vein, the problem of face detection and clustering is addressed jointly by~\cite{erdosclustering}, where a link-based clustering algorithm based on rank-1 counts verification merges frames based on a learned threshold.

Among recent works, \cite{tapaswi2019bcl} aim to estimate the number of clusters and their assignment and learn an embedding space that creates a fixed-radius balls for each character.
Datta~\etal~\cite{bestfgr2018} use video constraints to generate a set of positive and negative face pairs.
Perhaps most related to this work, \cite{ssiam} proposes two approaches:
\emph{SSiam}, a distance matrix between features on a subset of frames is used to generate hard positive/negative face pairs; and
\emph{TSiam} uses temporal constraints and mines negative pairs for singleton tracks by exploiting track-level distances.
Features are fine-tuned using a Siamese network with contrastive loss~\cite{contrastive_loss}.

Different from related work, we propose a simple, yet effective approach where weak labels are generated using a clustering algorithm that is independent of video/track level constraints to learn discriminative face representations.
In particular while \cite{merge-or-not,tapaswi2019bcl} require ground-truth labels to learn embeddings, CCL does not.

\vspace{2mm}
\noindent\textbf{Person identification}
is a related field of naming characters, and often employs \textbf{multimodal} sources of constraints/supervision such as:
clothing~\cite{tapaswi2012}, gender~\cite{mcafc}, context~\cite{zhang2013unified}, multispectral information~\cite{sharmams1};
audio including speech~\cite{paul2014conditional} and voice models~\cite{nagrani2017};
and textual cues such as weak labels obtained by aligning transcripts with subtitles~\cite{baeuml2013,everingham2006},
joint action/actor labels~\cite{miech2017learning} via transcripts, or
name mentions in subtitles~\cite{Haurilet2016}.
In contrast to the above, CCL is free from additional context and operates directly on detected faces (even without tracking).

\vspace{2mm}
\noindent\textbf{FINCH clustering algorithm.}
% Recently, advances in clustering approaches have contributed to better video face identification and clustering.
Sarfraz~\etal~\cite{finch} propose a new clustering algorithm (FINCH) based on first neighbor relations.
FINCH does not require training and relies on good representations to discover a hierarchy of partitions.
While FINCH demonstrates that pseudo labels can be used to train simple classifiers with cross-entropy loss~\cite{finch}, such a setup is challenging to use in the context of faces.
When using a partition at high purity with many more clusters than classes, faces of one character belong to multiple clusters, and the CE loss pushes these clusters and their samples apart -- an undesirable effect for our goal of clustering.
% the same character will be considered as belonging to dissimilar classes as it belongs to multiple clusters.
% being assigned to different clusters will be considered dissimilar classes, and thus the samples will be pushed away even if they belong to the same class.
Thus, in our work, we use FINCH to generate weak positive and negative face pairs that are used to improve video face clustering.
%\cite{merge-or-not} uses inverse reinforcement learning on a ground-truth dataset to find a reward function for deciding whether to merge a given pair of facial features.
%Similarly, \cite{tapaswi2019bcl} proposes to learn an embedding space that creates a fixed-radius ball for each character thus allowing to estimate the number of clusters.
%However, in both these works~\cite{merge-or-not,tapaswi2019bcl}, ground-truth labels are required to provide supervision to train the models.
%In contrast, in CCL, we learn the embedding in a self-supervised setting.
% Finally, there are some works that mine self-supervised labels from unlabeled sources which are similar in spirit to CCL.
% Fernando~\etal~\cite{fernando2017self} and Mishra~\etal~\cite{ref2}  use video frames reordering to generate positive or negative pairs; Wang~\etal~\cite{ref3} collect positive and negative labels from bounding boxes tracks.

\vspace{2mm}
\noindent\textbf{Clustering based learning.}
Recently clustering is also used to learn an embedding of data~\cite{caron2018deep,guo2017improved,jiang2016variational,xie2016unsupervised,yang2016joint}.
Almost all of these methods cluster data samples at each forward pass (or at pre-determined epochs) into a given number of clusters.
Subsequently, generated cluster labels or a loss over the clustering function are used to train a deep model (usually an auto-encoder).
These methods require to specify the number of clusters, that is often the target number of classes.
This is different from CCL that leverages a large number of small and pure clusters without specifying the particular number of clusters.

%{\color{red}our rebuttal differences to related work is so much better. this needs to be clearly written in multiple places:}
\vspace{2mm}
\noindent In summary, we show how pseudo-labels from clustering can be used effectively:
(i) without needing to know the number of clusters, and
(ii) by creating positive/negative pairs at a high-purity stage of clustering.

\section{Clustering based Representation Learning}
\label{sec:model}

\begin{figure*}[t]
\centering
\includegraphics[width=0.85\linewidth]{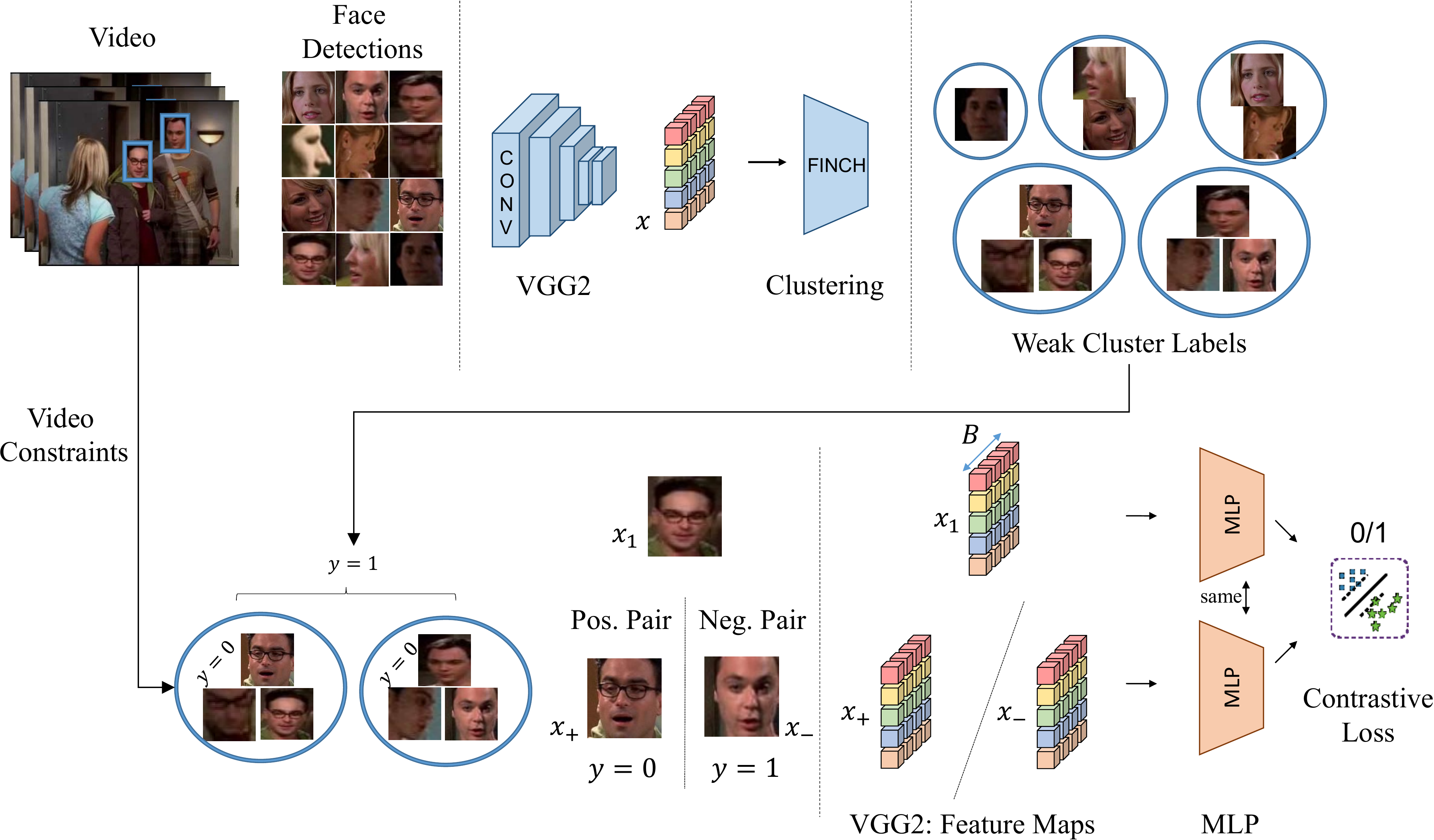}
\vspace{-0.2cm}
\caption{{\bf CCL training overview:}
Given a video with several face detections (top left), we first start by extracting features using a deep CNN and perform clustering using FINCH to obtain a large number of small but highly pure clusters (top).
We create several positive and negative face image pairs using these cluster labels and train an MLP to further improve the feature representation using the contrative loss (bottom).
At test time, the MLP is used as an embedding, and we cluster our samples using Hierarchical Agglomerative Clustering (HAC).}
\label{fig:pipeline}
\vspace{-0.5cm}
\end{figure*}

We present our method to improve face representations using weak cluster labels instead of manual annotations.
%In contrast, fine-tuning the original CNN typically requires supervised class labels. Thus, our approach has three key benefits: (i) it is easily applicable to new videos; (ii) it does not need large amounts of training data (few hundred tracks are enough); and (iii) specialized networks can be trained on each episode, film, or series.
We start this section by introducing the notation used throughout the remainder of the paper.
We then present a short overview and motivation for FINCH, followed by a description on how we use it to obtain automatic labels.
Finally, we discuss how we train our embedding and perform inference at test time.
Algorithm~\ref{algo:algo_overview} provides an overview of our proposed CCL approach, independent to the face clustering task.

\vspace{2mm}
\subsection{Preliminaries}
Let the dataset contain $N$ face tracks  $\mathcal{D} = \{(T_{i},y_{i})_{i=1}^{N}\}$, where each track $T_{i}$ has $K$, a variable number of face images, \ie~$T_{i} =\{(f_i^k)_{k=1}^K$\}.
$y_{i} \in \{1,\dots,C\}$ is the label (person identity) corresponding to the track $T_i$, where $C$ is the total number of characters in the video.
Please note that tracks are not necessary for the proposed learning approach, however, they are used during a part of the evaluation.

We use a CNN trained for recognizing faces (VGG2Face~\cite{vggface2}) and extract features for each face image in the track from the penultimate layer of the network.
The features for a track are vectors $\{\bx_i^{1},\ldots,\bx_i^{K}\}$ of size $\bx_i^k \in \mathbb{R}^{D \times 1}$, where $D$ denotes the feature dimension of the CNN feature maps.
We refer to these as \emph{base} features as they are not refined further by a learning approach.
% We use only center crop.

Track-level representations are formed by an aggregation function $\Phi: \{\bx_i^{1},\ldots,\bx_i^{K}\} \rightarrow \bv_i$, that combines $K$ features to produce a single vector $\bv_i \in \mathbb{R}^{D \times 1}$.
While there are several methods of aggregation such as Fisher Vectors~\cite{vf2}, Covariance Learning~\cite{wang2012cdl}, Quality Aware Networks~\cite{liu2017qan} or Neural Aggregation Networks~\cite{yang2017nan}, Temporal 3D Convolution~\cite{t3d}, we find that simple average pooling often works quite well.
In our case, averaging allows us to learn from features at the frame level (without requiring tracking), but evaluate at the track level.
% $\Phi$ allows us to aggregate track features into a compact and robust track-level representation.
% $\Phi$ computes track-level aggregated features by average pooling across the embedded frame-level representations~\cite{sharma2017}.

We additionally $\ell_{2}$ normalize the features to be unit vectors before using them for clustering, \ie~$\|\bv_i\|_2 = 1$.
Several previous works in this area~\cite{sharma2017,ssiam,tapaswi2019bcl,imptriplet,jfac} have used Hierarchical Agglomerative Clustering (HAC) as the preferred technique for clustering.
For a fair comparison, we too use HAC with the stopping condition set to the number of clusters equalling the number of main cast ($C$) in the video.
We use the minimum variance ward linkage~\cite{ward_linkage} for all methods.

\subsection{Clustering Algorithm}
Integral to our core method, we adopt the recently proposed FINCH algorithm~\cite{finch} to obtain weak labels from clustering.
FINCH belongs to the family of hierarchical clustering algorithms and automatically discovers groupings in the data without requiring hyper-parameters or a priori knowledge such as number of clusters.
The output of FINCH is a \emph{small} set of partitions that provide a fine to coarse view on the discovered clustering.
Note that this is different from classical HAC methods where each iteration merges one sample with existing clusters providing a complete list of partitions ranging from $N$ clusters (all samples in their own cluster) to 1 cluster (all samples in one cluster).

FINCH works by linking first neighbors of each sample.
An adjacency matrix is defined between all sample pairs:
\begin{equation}
A(i,j) =
\begin{dcases*}
   1                         & if $j= \kappa_i^1$ or $\kappa_j^1=i$ or $\kappa_i^1=\kappa_j^1$ \\
   0                         & otherwise \, ,
\end{dcases*} \\
\label{eq:finch}
\end{equation}
where $\kappa_i^1$ represents the first neighbor of sample $i$.
The adjacency matrix joins each sample $i$ to their first neighbors via $j=\kappa_i^1$, enforces symmetry through $\kappa_j^1=i$, and links samples that share a common first  neighbor $\kappa_i^1=\kappa_j^1$.
Clustering is performed by identifying the connected components in the adjacency matrix.

\vspace{1mm}
\noindent\textbf{Why use FINCH?}
% \makarand{I have updated this part a lot, can you check if its ok? we weren't really talking about why use FINCH earlier} \vivek{Yes, it looks good!}
%{\color{red}should this be \cite{finch}?}
% There are two differences from previous methods~\cite{finch}
% that have used clustering labels as a form of category labels:
% (i) we achieve a very high purity (usually over 99\%) within each cluster; and
% (ii) the number of clusters is much greater than the true number of categories, and each cluster typically contains few samples ($< 10$).
FINCH is a suitable clustering algorithm for our task for several reasons:
(i) it does not require specifying the number of clusters;
(ii) it provides clusters with very high purity at early partitions; and
(iii) it is a fast and scalable algorithm.

As demonstrated in~\cite{finch}, FINCH outperforms Affinity Propagation~(AP), Jarvis and Patrick~(JP), Robust Continuous Clustering~(RCC), and Rank Order Clustering~(RO) on several datasets.
Additionally, all algorithms AP, JP, RCC, and RO (except FINCH) have memory requirements quadratic in number of samples and are unable to cluster features on our machines with 128 GB RAM. 

The first partition of FINCH corresponds to linking samples through the first neighbor relations, while the second partition links clusters created in the first step.
We use clusters from the second partition of the FINCH algorithm to mine positive and negative pairs as this partition increases diversity, without compromising on quality of the labels.
We find that most samples clustered in the first partition are from within a track (\eg~neighboring frames), while those in the second partition are often from across tracks (\cf~Fig.~\ref{fig:hist} (right)).
Empirically, we analyze the impact of choosing different partitions through ablation studies.

\vspace{1mm}
\noindent\textbf{Using FINCH to obtain weak labels.}
We treat each face image as a sample and perform clustering with FINCH to obtain the second clustering partition $P_2 = \{G_1, \ldots, G_M\}$.
$G_i$ corresponds to the $i^{\mathrm{th}}$ group of samples.
Based on the cluster labels, we obtain positive and negative pairs as:
\begin{itemize}
\item  \textbf{PosC}: All face images in a cluster are considered for positive pairs.
For clusters that have less than 10 samples, we also create positive pairs by combining samples from the current cluster $G_{p1}$ with randomly sampled frames from another cluster $G_{p2}$, where $G_{p2}$ is among the $Z = 25$ nearest clusters to $G_{p1}$.
\item \textbf{NegC}: We create negative pairs by combining samples from a cluster $G_{n1}$ with randomly sampled frames from a different cluster $G_{n2}$, where $G_{n2}$ is among the $Z$ farthest clusters to $G_{n1}$.
\end{itemize}
% \makarand{does Z depend on the number of clusters? in table 4, when you have 101 clusters at 4th partition, do you still use 25 closest/farthest clusters? that might explain the drop in performance no?} \vivek{Yes, when smaller than 25 clusters, we opt for farthest 20\% clusters. I guess, for 13 clusters, it was 3 farthest clusters.}

\vspace{1mm}
\noindent\textbf{$K$-means as a baseline.}
As an alternative to FINCH, we perform experiments with $K$-means as a baseline to obtain clusters that provide weak labels.
Note that $K$-means has an important disadvantage -- we need to know the number of clusters before hand.
For a fair empirical comparison, we use the number of clusters estimated by FINCH at each partition.
Specifically, we use \emph{MiniBatch $K$-means}~\cite{scully2010minibatchkmeans} that is more suitable to work with large datasets.
Our analysis shows that FINCH provides purer clusters that yield more accurate pairwise labels.

\subsection{Video Level Constraints}

Video face clustering often employs face tracking to link face detections made in a sequence of consecutive frames.
Note that tracking can be thought of as a form of clustering in a small temporal window (typically within a shot).
Using tracks, positive pairs are created by sampling frames within a track, while negative pairs are created by analyzing co-occurring tracks~\cite{cinbis2011unsupervised,bestfgr2018,tc,wu2013constrained}.

In this work, we consider video constraints that can be derived at a frame level (\ie~without tracking).
We include co-occurring face images appearing in the same frame as potential negative pairs (\textbf{NVid}).
About 50\%-70\% of the face images appear as singletons in a frame (as shown by TSiam~\cite{ssiam}), and do not yield negative pairs through the video level constraints.
Thus, being able to sample negative pairs using the cluster labels as discussed above is beneficial.

In addition to sampling negative pairs, we also use co-occurrence to correct clusters provided by FINCH.
For example, if a cluster contains samples from a known negative pair, we keep the sample closest to the cluster mean, and create a new cluster using the rejected sample.

% Given the assignment of faces to weak clusters, we would like to generate positive and negative face pairs based on sample assignment to cluster driven by video constraints.
% We consider three rules:
% \begin{itemize}
    % \item \makarand{i don't understand this. do you change how the adjacency links based on co-occurring faces? what does the we make a new cluster part mean?} No co-occurring faces in the same frame are assigned to the same cluster. We make a new cluster with the rejected sample (one for each co-occurring face) and the most similar sample stays wrt. the average representation of the cluster. 
    % \item For frames with singleton (no co-occurring) face, we incorporate negative cluster pairs by randomly sampling frames from the farthest $25$ clusters. Similar to~\cite{ssiam} handles negative pairs for singleton (not co-occurring) too, but we do it at frame/cluster level, while they do it for track-level supervision. For BBT-0101 we have 50\% frames with no co-occurring faces, while this can be large as 70\% for BF-0502.
    % \item  All faces in a given cluster is considered positive pairs. For clusters with less than 10 faces, we incorporate positive cluster pairs by randomly sampling frames from the closest $25$ clusters.
% \end{itemize}

\subsection{Training and Inference}
We use the weak cluster labels and video constraints to obtain positive and negative face pairs for training a shallow Siamese network.
We adopt the Contrastive loss~\cite{contrastive_loss} as it brings together positive samples, in contrast to the widely used Triplet loss~\cite{facenet} that only maintains a margin between positive and negative samples.
We sample a pair of face representations $\bx_1$ and $\bx_2$ with $y = 0$ for a positive pair and $y = 1$ when corresponding to a negative pair and train a multi-layer perceptron $Q_\theta$ by minimizing
\begin{equation}
\label{eq:contrasiveloss}
\begin{split}
\mathcal{L} \left(W, y, Q_\theta(\bx_1), Q_\theta(\bx_2) \right) =  \qquad \qquad \qquad \qquad \qquad \\
\frac{1}{2} \left( (1 - y) \cdot (d_W)^2 + y \cdot (\max(0, m - d_W))^2 \right) \, ,
\end{split}
\end{equation}
where $W \in \mathbb{R}^{D \times d}$ is a linear layer that embeds $Q_\theta(\cdot)$ such that $d \ll D$ (in our case, $d = 2$).
Here, each face image is encoded as $Q_\theta(\bx)$, where $\theta$ corresponds to the trainable parameters.
We find that $Q_\theta(\cdot)$ performs well when using a single linear layer.
$d_W$ is the Euclidean distance $d_W = \| W \cdot Q_\theta(\bx_1) - W \cdot Q_\theta(\bx_2) \|^2$, and
$m$ is the margin, empirically chosen to be 1.
Fig.~\ref{fig:pipeline} illustrates this learning procedure.

During inference, we compute embeddings for face images using the MLP trained above.
For frame-level evaluation, we cluster the features for each face image, while for track-level evaluation, we first aggregate image features using mean pool, followed by HAC.
We report the clustering performance at the ground truth number of clusters (\ie~the number of main characters in the video).
% Algorithm~\ref{algo:algo_overview} provides an overview of the complete CCL procedure.

\begin{algorithm}[t]
\caption{Overview of CCL (task agnostic).
% Mining positive and negative pairs via clustering and video constraints.
}\label{algo:algo_overview}
\begin{algorithmic}
\STATE \textbf{Input:}
Feature maps $X = \{(\bx_n)_{n=1}^N\} \in \mathbb{R}^D$. \\
$N$ is total number of samples.
$D$ is feature dimension.
\STATE \textbf{Output:} Final clustering of the $N$ samples. 
\STATE \textbf{Procedure:}
\STATE \textbf{1.}
Compute the partitions using FINCH clustering algorithm.
$P_{1:L} = \mbox{FINCH}(X)$.
\STATE \textbf{2.}
Select second partition $P_2 = \{G_1, \ldots, G_M\}$.
\STATE \textbf{3.}
Train model parameters for 20 epochs:
\STATE \hspace{5mm} -- Sample positive and negative pairs for a batch.\\
$(\bx_{p1}, \bx_{p2}) \sim (G_{p1}, G_{p2})$ and
$(\bx_{n1}, \bx_{n2}) \sim (G_{n1}, G_{n2})$.
\STATE \hspace{5mm} -- Train MLP $Q_\theta(\cdot)$ using Contrastive loss.
\STATE \textbf{4.}
Compute embeddings for each sample using $Q_\theta(\bx_n)$ and perform HAC to get $C$ clusters.
\end{algorithmic}
\end{algorithm}

\subsection{Implementation Details}
\label{subsec:implementation}

% \vspace{2mm}
\noindent\textbf{CNN.}
We employ the VGG-2 Face CNN~\cite{vggface2} (ResNet 50) pre-trained on MS-Celeb-1M~\cite{msceleb1m} and fine-tuned on 3.31M face images of 9131 subjects (VGG2 data).
% We use the VGG-2 face ResNet50 model.
We extract \texttt{pool5\_7x7\_s1} features by first resizing RGB face images to $224 \times 224$ and pushing them through the CNN, resulting in $\bx^i_k \in \mathbb{R}^{2048}$.
At test time, we compute face embeddings using the learned MLP $Q_\theta$ and $\ell_{2}$ normalize before HAC.

\vspace{1mm}
\noindent\textbf{Siamese network MLP.}
The network comprises of fully-connected layers: $\mathbb{R}^{2048} \rightarrow \mathbb{R}^{256} \rightarrow \mathbb{R}^{2}$.
Note that the second linear layer is part of the contrastive loss (corresponds to $W$ in Eq.~\ref{eq:contrasiveloss}), and we use the feature representations at $\mathbb{R}^{256}$ for clustering.
The model is trained using the Contrastive loss, and parameters are updated using Adam optimizer.
We initialize the learning rate with $10^{-5}$ and decrease it by a factor of 10 at 15 epochs.
We train the model for 20 epochs. We use batch normalization.

\vspace{1mm}
\noindent\textbf{Sampling positive and negative pairs for training.}
An epoch corresponds to sampling pairs from all clusters created by the FINCH algorithm.
We consider 5 clusters per batch.
For each data point in a cluster, we randomly sample from the same (or nearest) cluster and create one positive pair, and sample from the farthest clusters to create two negative pairs.
Finally, we randomly subsample the above list to obtain 25 positive and 25 negative pairs for each cluster, resulting in a batch with 250 pairs (125 positive/negative).

\section{Evaluation} \label{sec:eval}
We present our evaluation on three challenging datasets.
We first describe the clustering metric, followed by a thorough analysis of the proposed methods, and end this section with a comparison of our approach against state-of-the-art.

\begin{table}[t]
\small
\tabcolsep=1.2mm
\begin{center}
\caption{Dataset statistics for BBT~\cite{wu2013simultaneous,wu2013constrained,imptriplet}, BF~\cite{jfac} and ACCIO~\cite{jfac}.}
% We show the number of characters, tracks and faces, and the ratio of largest to smallest cluster size.} 
\label{table:stats}
% \resizebox{\linewidth}{!}{
\begin{tabular}{lcccc}
\toprule
				&		 &   This work  &               & Previous work \\
Datasets        & \#Cast &  \#TR (\#FR) &    LC/SC (\%) & \#TR (\#FR)\\ 
\midrule
BBT-0101 		& 	5	&	644 (41220)	&	37.2 / 4.1  & 182 (11525)\\
BF-0502 			&	6	&	568 (39263)	&	36.2 / 5.0  & 229 (17337)\\
%NH            	&	5	&	240 (16872)	& 	43.1 / 7.4  & 76 (4660)~\cite{wu2013constrained,jfac}\\
ACCIO	   		&	36	&	3243 (166885) & 30.93/0.05&	3243 (166885) \\
\bottomrule
\end{tabular}
\end{center}
\vspace{-0.7cm}
\end{table}

\vspace{2mm}
\noindent\textbf{Datasets.}
% We present a summary of the dataset used in this work in Table~\ref{table:stats}.
We conduct experiments on three popular video face clustering datasets:
(i) \textit{Buffy the Vampire Slayer} (BF)~\cite{baeuml2013,jfac} (season 5, episodes 1): a drama series with several dark scenes and non-frontal faces;
(ii) \textit{Big Bang Theory} (BBT)~\cite{baeuml2013,tapaswi2012,wu2013simultaneous,imptriplet} (season 1, episodes 1): a primarily indoors sitcom with a small cast, and
(iii) \textit{ACCIO-1}~\cite{accio}: the first movie in the  ``\textit{Harry Potter}'' series featuring a large number of scenes at night and a higher number of characters.

We follow the protocol used in several recent video face clustering works~\cite{cinbis2011unsupervised,ssiam,wu2013constrained,imptriplet,jfac} that focus on improving feature representations for video-face clustering.
First, we assume the number of main characters is known.
Second, as the labels are obtained automatically, we learn episode specific embeddings.
% and train on BF episode 2, BBT episode 1, and ACCIO.
We use the tracks provided by~\cite{ssiam}
% which is an updated version of face tracks from~\cite{baeuml2013} 
that incorporate several detectors to encompass all pan angles and in-plane rotations up to 45 degrees.
The face tracks are obtained by an online tracker that uses a particle filter.

Table~\ref{table:stats} presents key information about the datasets such as the number of tracks (\#TR) and face images (\#FR).
In particular, note that previous works (except~\cite{ssiam}) use much smaller datasets.
% In particular, we indicate the number of tracks (\#TR) and frames (\#FR)  used in our and previous works, 
It is also important to note that different characters have wide variations in the number of tracks, indicated by the cluster skew between the largest class (LC) to the smallest class (SC).

\vspace{2mm}
\noindent\textbf{Evaluation metric.}
We use Clustering Accuracy (ACC)~\cite{ssiam} also called Weighted Clustering Purity (WCP)~\cite{tc} as the metric to evaluate the quality of clustering.
Accuracy is computed by assigning the most common ground truth label within a cluster to all elements in that cluster:
\begin{equation}
\mbox{ACC} = \frac{1}{N} \sum_{c = 1}^{|C|} n_c \cdot p_c \, ,
\end{equation}
where $N$ is the total number of samples,
$n_c$ is the number of samples in the cluster $c$, and
cluster purity $p_c$ is measured as the fraction of the largest number of samples from the same label to $n_{c}$.
$C$ corresponds to the number of main casts.
For ACCIO, we also report B-Cubed Precision (P), Recall (R) and F-measure (F)~\cite{amigo2009comparison}.
Unless stated otherwise, clustering evaluation is performed at track-level.

\subsection{Clustering Performance and Generalization}

\begin{table}[t]
\small
\tabcolsep=0.1cm
\centering
\caption{Clustering accuracy at track-level.
Comparison against all evaluated models.}
%\vspace{-3mm}
\label{table:comparison} 
\begin{tabular}{lccccc}
\toprule 
Train/Test 	& Base  	& PRF~\cite{psuedo1}  & TSiam~\cite{ssiam}     &  SSiam~\cite{ssiam}  &  CCL	   \\
\midrule
BBT-0101    & 0.932 	& 0.930      & 0.964 	& 0.962	&  \textbf{0.982} 	  \\
BF-0502     & 0.836 	& 0.814      & 0.893 	& 0.909 &  \textbf{0.921} 	  \\
\bottomrule
\end{tabular}
%\vspace{-3mm}
\end{table}

\begin{table}[t]
{\small
\tabcolsep=0.12cm
\caption{Clustering accuracy at track-level, extending to all named characters.
BBT-0101 has 5 main and 6 named characters;
BF-0502 has 6 main and 12 named characters.}
%\vspace{-3mm}
\label{table:generalization_more_characters} 
\begin{center}
% \resizebox{8.5cm}{!}{
\begin{tabular}{l|ccc|ccc}
\toprule
    & \multicolumn{3}{c}{BBT-0101}  & \multicolumn{3}{|c}{BF-0502} \\
    %\cline{2-5}
     	 & TSiam & SSiam & CCL       	  & TSiam & SSiam & CCL        \\
\midrule
Main cast  		& 0.964 & 0.962   &  \textbf{0.982}   	 & 0.893 & 0.909   &  \textbf{0.921} \\
All named       & 0.958 & 0.922  &  \textbf{0.966}	  & 0.829 & 0.870  &  \textbf{0.903}    \\
\bottomrule
\end{tabular}
\end{center}}
\vspace{-0.7cm}
\end{table}

\begin{table*}[ht]
{\small
\tabcolsep=0.05cm
\begin{center}
\caption{A study on the impact of clustering algorithm.
FINCH partitions are created for each dataset, and shown as separate table rows.
In each row, results are presented for FINCH (above) and MiniBatch $K$-means (below).
From left to right, \textbf{Part.} indicates the partition level of FINCH.
\textbf{\#C} is the total number of clusters in that partition as estimated by FINCH.
% Each row is represented as ($_{Mini-Batch KMeans}^{Finch}$) for the \#C.
Largest and smallest cluster sizes are indicated as \textbf{LC/SC}.
Clustering purity of FINCH/$K$-means clusters (before CCL) is presented as \textbf{ACC}.
\textbf{L+/L-} represents the number of samples correctly and wrongly clustered for the given partition.
Finally, \textbf{CCL-ACC} is the performance of CCL: by training a model using weak labels from FINCH/$K$-means estimated clusters.
% CCL when trained with Finch estimated weak labels and evaluated on groundtruth clusters.
% Weighted Clustering Purity~(ACC) or Clustering Accuracy~(\%) performance comparison  over all three dataset BBT-0101, BF-0502 and ACCIO.
% ACC is base performance before training CCL, and LC/SC denotes largest class (LC) / smallest class (SC) is the class balance percent of the given partition and L+/L- is the number of samples correctly and wrongly clustered for the given partition. $-$ means CCL was not trained for these partitions.
} 
\label{table:partitions} 
%\begin{adjustbox}{angle=90}
\resizebox{17.5cm}{!}{
\begin{tabular}{c|ccccc|ccccc|ccccc}%ccccc}
\toprule
FINCH & \multicolumn{5}{c|}{BBT-0101} & \multicolumn{5}{c|}{BF-0502} & \multicolumn{5}{c}{ACCIO} \\
Partition & \#C & LC/SC  & ACC & L+/L- & CCL-ACC
      & \#C & LC/SC  & ACC & L+/L- & CCL-ACC
      & \#C & LC/SC  & ACC & L+/L- & CCL-ACC \\
% \midrule
   & \textbf{41220} &  &  & & @\#C=5 & \textbf{39263} &  &  &  & @\#C=6 & \textbf{166885} &  &  && @\#C=36\\ 
% \midrule
\midrule
\multirow{2}{*}{1} & \multirow{2}{*}{10156}
    &48/2&0.997&41128/92&0.978&   \multirow{2}{*}{9677} &42/2  &0.994&39054/209&0.915 & \multirow{2}{*}{21444} & 3937/2&0.847&141496/25389&0.816  \\ 
    &&1030/1&0.988&40744/476&0.971& & 758/1 &0.986&38717/546&0.909 && 16041/1&0.899&150177/16708&0.841   \\ 
\midrule
\multirow{2}{*}{\textbf{2}} & \multirow{2}{*}{\textbf{2236}}
    &777/4&0.990&40809/411&\textbf{0.995}&   \multirow{2}{*}{\textbf{2167}} &1346/4  &0.978&38414/849&\textbf{0.938} & \multirow{2}{*}{\textbf{3972}} & 33461/4&0.832&138903/27982&\textbf{0.837}  \\ 
    &&1236/1&0.987&40687/533&0.991& & 1566/1 &0.971&38161/1102&0.923 && 18076/1&0.866&144613/22272&\textbf{0.857}   \\ 
\midrule
\multirow{2}{*}{3} & \multirow{2}{*}{490}
    &2031/8&0.974&40149/1071&0.954&   \multirow{2}{*}{560} &2574/9  &0.957&37588/1675&0.922 & \multirow{2}{*}{944} & 41405/13&0.812&135543/31342&0.801  \\ 
    &&1568/1&0.979&40367/853&0.957& & 2120/1 &0.962&37806/1457&0.926&& 9838/1&0.845&141036/25849&0.813   \\ 
\midrule
\multirow{2}{*}{4} & \multirow{2}{*}{101}
    &7258/32&0.968&39936/1284&0.943&   \multirow{2}{*}{127} &9335/27  &0.923&36260/3003&0.903 & \multirow{2}{*}{161} & 70825/39&0.784&130987/35898&0.767  \\ 
    &&2019/1&0.981&40475/745&0.945& & 2360/1 &0.956&37555/1708&0.923 & & 16502/1&0.800&133639/33246&0.782   \\ 
\midrule
\multirow{2}{*}{5} & \multirow{2}{*}{13}
    &10.9K/0.K&0.968&39.9K/1.2K&0.943&   \multirow{2}{*}{24} &13.7K/0.2K  &0.895&35.1K/4.0K&0.869 & \multirow{2}{*}{24} & 84.8K/0.2K&0.690&115.1K/51.7K&0.652  \\ 
    &&7.3K/1&0.966&39.8K/1.3K&0.931 & & 3.7K/0.2K &0.920&36.1K/3.1K&0.875 & & 26.7K/1&0.724&120.9K/45.9K&0.696   \\ 
%\midrule
%\multirow{2}{*}{6} & \multirow{2}{*}{2}
%    &26.4K/14.7K&0.724&29.8K/11.3K&-&   \multirow{2}{*}{5} & 20.8K/1.2K  &0.799&31.3K/7.8K&- & \multirow{2}{*}{4} & 152.4K/1.1K&0.387&64.6K/102.2K&-  \\ 
%    &&26.5K/14.7K&0.727&29.9K/11.2K&  -& & 16.7K/3.3K &0.910&35.7K/3.5K&- & & 60.8K/17.2K&0.506&84.4K/82.4K&-   \\ 
%\midrule
%\multirow{2}{*}{7} & \multirow{2}{*}{1}
%    &41.2K&0.372&15.3K/25.8K&-&   \multirow{2}{*}{1} &
%    39.3K&0.361&14.1K/25.0K&- & \multirow{2}{*}{1} &
%    166.9K&0.309&51.6K/115.2K&-  \\ 
%    &&41.2K&0.372&15.3K/25.8K&-&&
%    39.3K&0.361&14.1K/25.0K&-&&
%    166.9K&0.309&51.6K/115.2K&-   \\ 
\bottomrule
\end{tabular}}
%\end{adjustbox}
\end{center}}
\vspace{-0.6cm}
\end{table*}

%%%%%%%%%%%%%%%%%%%%%%%%%%%%%%%%%%%%%%%%%%%%%%%%%%%%%%%%%%%%%%%%
% Comparing models on training episodes
%%%%%%%%%%%%%%%%%%%%%%%%%%%%%%%%%%%%%%%%%%%%%%%%%%%%%%%%%%%%%%%%

\vspace{2mm}
\noindent\textbf{Comparison against baselines.}
We present a thorough comparison of CCL against related baselines that employ a similar learning scheme.

In pseudo-relevance feedback~\cite{psuedo1,pseudo2} (PRF), samples are treated independent of each other, and pairs are created by considering closest positives and farthest negatives.
However, these pairs often provide negligible training signal (gradients) as they already conform to the loss requirements.

SSiam~\cite{ssiam} is an improved version of PRF with batch processing where farthest positives and closest negatives are formed by looking at a batch of queries rather than whole dataset.
This creates harder training samples that show improved performance over PRF.

In TSiam~\cite{ssiam}, positive pairs are formed by looking at samples within a track (a track is treated as a cluster of face images) and negative pairs by using co-occurrence and distances between track representations.

Finally, our proposed method CCL does not rely on tracking and uses pure clusters created by an automatic partitioning method (FINCH).
We sort clusters by distances between their mean representations and then form positive and negative pairs.
Table~\ref{table:comparison} shows that CCL outperforms both SSiam and TSiam, and also provides significant gains over the base VGG2 features on BBT and BF.

\vspace{2mm}
\noindent\textbf{Generalization to unseen characters.}
We further study how CCL can cope with adding unseen characters at test time by clustering all named characters in an episode.
Results from Table~\ref{table:generalization_more_characters} show that CCL is better poised at clustering new characters and achieves higher performance in all cases.
We believe that CCL's performance scales well as it is trained on a diverse set of pairs and can generalize to unseen characters.

%Fig.~\ref{fig:hist} (right) shows the number of faces in each cluster and the number of tracks that they are derived from.
%CCL's ability to obtain positive pairs from faces across multiple tracks (different from TSiam) might be an explanation for improved performance.
%This means that our learning approach sees relatively hard positive pairs, and standard (neither hard nor easy) negative pairs.
%We believe that together with the contrastive loss that brings together samples from positive pairs, this is a good way to learn representations, and is an important factor for the increased performance.

\subsection{Sources of Positive and Negative Pairs}
\begin{table}[t]
\tabcolsep=0.2cm
\small
\centering
\caption{Impact of mining positive and negative pairs from different sources. Track-level accuracy of CCL.}
\label{table:ablation_pn_clusters}
%\vspace{-0.30cm}
\begin{tabular}{llccccc}
\toprule
 & & PosC & NegC & NVid & BBT-0101 & BF-0502 \\
\midrule
1 & Base & - & - & - & 0.932 & 0.836 \\
\midrule
2 & & \ding{51} & - & - & 0.951 & 0.859 \\
3 & & - & \ding{51} & - & 0.968 & 0.883 \\
4 & & \ding{51} & - & \ding{51} & 0.956 & 0.865 \\
5 & & \ding{51} & \ding{51} & - & 0.971 & 0.896 \\
6 & & - & \ding{51} & \ding{51} & 0.979 & 0.917 \\
\midrule
7 & CCL & \ding{51} & \ding{51} & \ding{51} & \textbf{0.982} & \textbf{0.921} \\
\bottomrule
\end{tabular}
\vspace{-0.6cm}
\end{table}
In Table~\ref{table:ablation_pn_clusters}, we present the importance of each source from which we obtain positive and negative pairs.
Recall that positive pairs obtained from clusters are denoted as \textbf{PosC};
negative pairs from clusters as \textbf{NegC}; and
negative pairs using video constraints as \textbf{NVid}.

Rows 2 to 6 evaluate different combinations of the sources of pairs and show their relative importance.
It can be seen that negative pairs alone (row 3) are more important than positive pairs (row 2).
Additionally, using pairs from the clusters alone (row 5) provides performance similar to TSiam and SSiam (refer to Table~\ref{table:comparison}).
In Fig.~\ref{fig:hist}, we show the number of faces in each cluster and the number of tracks that they are derived from.
CCL's ability to obtain positive pairs from faces across multiple tracks (different from TSiam) might be an explanation for improved performance.
Further, including negatives from videos (row 7) provides best performance.
This means that our learning approach sees relatively hard positive pairs, and standard (neither hard nor easy) negative pairs.
We believe that together with the contrastive loss that brings together samples from positive pairs, this is a good way to learn representations, and is an important factor for the increased performance.

%Additionally, using pairs from the clusters alone (row 5) provides performance similar to TSiam and SSiam (refer to Table~\ref{table:comparison}), and including negatives from videos (row 7) provides best performance.
%{\color{red}need to re-think about how does this relates with the discussion above where we say posc is important, there are hard positives, but simple negatives}

\subsection{Impact of Clustering Algorithm}
\label{sec:ablation}
We now present a study on the impact of the clustering algorithm used for obtaining weak labels.
We first obtain the hierarchy of partitions by processing each dataset with the FINCH algorithm.
Then, as a comparison, we use MiniBatch $K$-means with $K$ set to the number of clusters provided by FINCH.
Each row of Table~\ref{table:partitions} shows performance at one partition.
Results using FINCH are in the top half and $K$-means in the bottom half.
The number of clusters in higher partitions (6 onwards) are less than the number of characters and are omitted.
% Specifically, we use MiniBatch $K$-means for this evaluation as computing the full distance matrix would require $\sim$120 GB memory on larger datasets such as ACCIO.

\begin{figure}[t]
\centering
\includegraphics[width=0.48\linewidth]{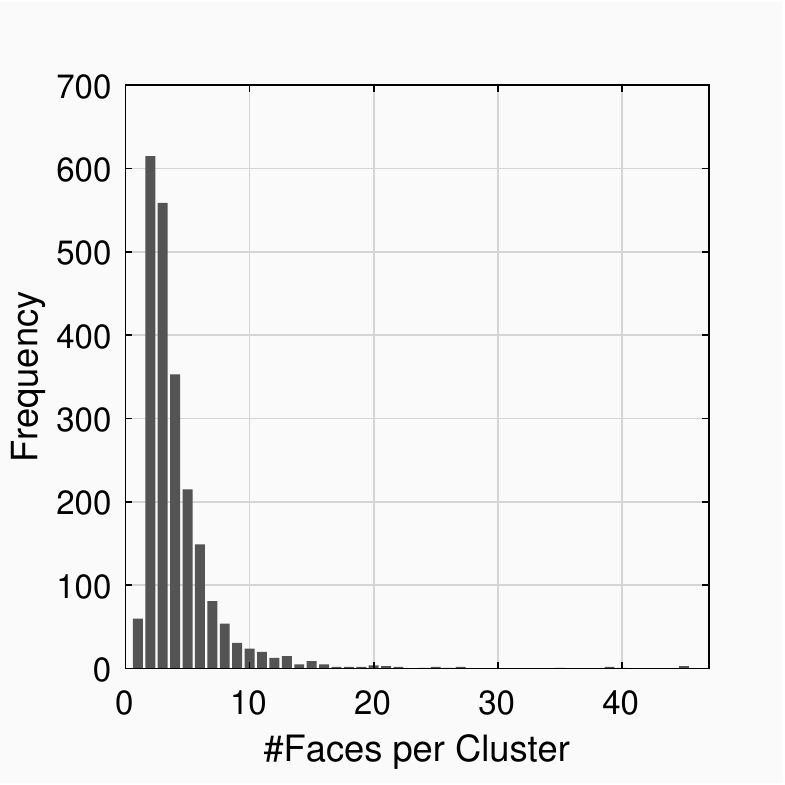} \hfill
\includegraphics[width=0.48\columnwidth]{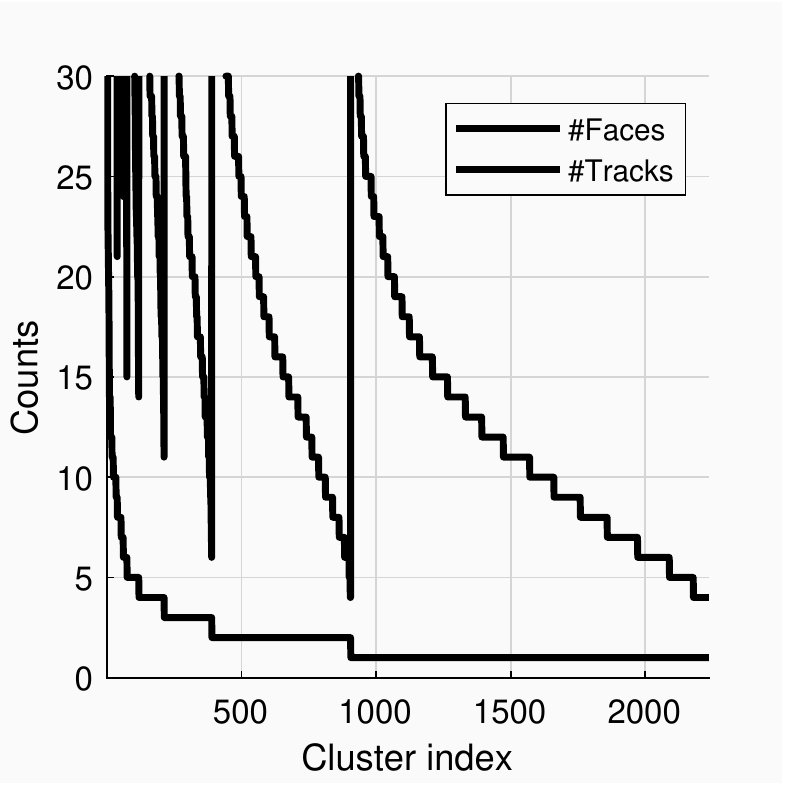}
\vspace{-0.3cm}
\caption{Key characteristics of FINCH (second partition) clustering for BBT-0101.
\textbf{Left:} Histogram showing the number of faces in a cluster. Even if most clusters have less than 5 samples, we are able to obtain meaningful positive and negative pairs to train our model.
\textbf{Right:} We plot the number of faces and number of tracks for each of the cluster indices (sorted by size for convenience).
About 900 of the 2200 clusters created by FINCH contain faces from more than one track, leading to increased diversity of pairs.}
\label{fig:hist}
\end{figure}

We observe that FINCH not only automatically discovers meaningful partitions of the data but also achieves higher-performance (ACC in Table~\ref{table:partitions}) as compared to $K$-means, especially at higher number of clusters.
It is interesting to note the number of samples that are clustered correctly or wrongly (L+/L- in Table) as these play an important role while creating weak labels.
Fig.~\ref{fig:hist} (left) shows that while the clusters are often very small ($< 10$ samples), they contain have faces from more than one track (right).
% and wrongly play an important role - and with these weak labels which when used to train CCL and then evaluated on the groundtruth number of clusters clusters:
CCL with FINCH at the second partition obtains high performance after training with weak labels~(CCL-ACC) and outperforms labels provided by $K$-means clustering.

% Further note that, L+/L- in Table~\ref{table:partitions} shows the spread of clusters. In the first partition, the samples are well spread out to lots of clusters that implies that the clusters are pure but then they miss to incorporate a wide range of faces coming from tracks throughout the video. One can easily observe that in Fig.~\ref{fig:hist}~(Right) the spread of clusters combine multiple tracks in the second partition, this help to make pairs and learn discriminative face features for each identity.

Additionally, as indicated earlier, running FINCH on large datasets such as ACCIO takes $\sim$2 minutes with fast nearest neighbors as compared to MiniBatch $K$-means that takes $\sim$1 hour even though it is optimized for large datasets.

\subsection{Comparison with the state-of-the-art}

\begin{table}[t]
\tabcolsep=0.15cm
\begin{center}
\caption{Comparison to state-of-the-art with clustering accuracy (\%) at frame level.
Note that many previous works use fewer tracks (\# of frames) (also indicated in Table~\ref{table:stats}) making the task relatively easier.
We use the tracks from~\cite{finch,ssiam}.}
\label{table:soa}
\resizebox{.48\textwidth}{!}{
\begin{tabular}{lcc|cc}
\toprule
\multirow{2}{*}{Method} &
\multirow{2}{*}{BBT-0101} & \multirow{2}{*}{BF-0502}  &
						 \multicolumn{2}{c}{Data Source} \\
                          & & &  BBT & BF \\
\midrule
FINCH {\scriptsize(CVPR '19)}~\cite{finch}                                       & {99.16} & 92.73  & \cite{baeuml2013}*   & \cite{baeuml2013}* \\
\midrule
ULDML {\scriptsize(ICCV '11)}~\cite{cinbis2011unsupervised}  & 57.00  & 41.62    & $-$ & \cite{cinbis2011unsupervised} \\
HMRF {\scriptsize(CVPR '13)}~\cite{wu2013constrained}        & 59.61   & 50.30    & \cite{roth2012robust} & \cite{everingham2006}  \\
HMRF2 {\scriptsize(ICCV '13)}~\cite{wu2013simultaneous} 	 &  66.77  & $-$      & \cite{roth2012robust} & $-$ \\
WBSLRR {\scriptsize(ECCV '14)}~\cite{xiao2014weighted}       & 72.00  & 62.76    & $-$ & \cite{everingham2006} \\
VDF {\scriptsize(CVPR '17)}~\cite{sharma2017}       		& 89.62  & 87.46     & \cite{baeuml2013} &\cite{baeuml2013}\\
Imp-Triplet {\scriptsize(PacRim '16)}~\cite{imptriplet}      & 96.00  & $-$       & \cite{roth2012robust}  & $-$\\
JFAC {\scriptsize(ECCV '16)}~\cite{jfac}                     & $-$    & 92.13    & $-$ & \cite{everingham2006} \\

TSiam {\scriptsize(FG '19)}~\cite{ssiam}                                       & {98.58} & {92.46}  &\cite{baeuml2013}*  & \cite{baeuml2013}*   \\
SSiam {\scriptsize(FG '19)}~\cite{ssiam}                                       & {99.04} & 90.87  & \cite{baeuml2013}*   & \cite{baeuml2013}* \\
%\midrule
%Base & 94.00 & 91.20 &\cite{baeuml2013}*  & \cite{baeuml2013}*   \\
\midrule
%OSiam                                       & {96.11} & {85.59}  &\cite{ssiam}   & \cite{ssiam}    \\
\textbf{CCL} (Ours with HAC)                                     & \textbf{99.56} & \textbf{93.79}  &\cite{ssiam}   & \cite{ssiam}    \\
\bottomrule
\end{tabular}}
\end{center}
\vspace{-0.8cm}
\end{table}

\begin{table}[t]
\tabcolsep=0.15cm
\small
\centering
\caption{Frame-level accuracy (\%) of CCL and comparison to TSiam and SSiam over all datasets: BBT-0101, BF-0502 and ACCIO.}
\label{table:accuracy_all}
\vspace{-0.30cm}
\begin{tabular}{lccccc}
\toprule
 & \#Cast & Base & TSiam~\cite{ssiam} & SSiam~\cite{ssiam} & CCL\\
\midrule
BBT-0101  & 5 & 94.00	& 98.58 & 99.04 & \textbf{99.56} \\
BF-0502   &	6 & 91.20 & 92.46 & 90.87 & \textbf{93.79} \\
ACCIO     &	36& 79.90 & 81.30 & 82.00 & \textbf{83.40} \\
\bottomrule
\end{tabular}
%\vspace{-0.7cm}
\end{table}

\begin{table}[t]
\tabcolsep=0.20cm
\small
\begin{center}
\caption{Comparison of CCL with the state-of-the-art on ACCIO, evaluated at \textbf{36} and \textbf{40} clusters.
Clustering accuracy at frame-level.}
\label{table:accio}
\begin{tabular}{l|ccc}
\toprule
\multicolumn{4}{c}{\textbf{\#Clusters = 36}} \\
Methods & P & R & F   \\
\midrule
%IL-HC {\scriptsize(AAAI '18)}~\cite{merge-or-not} & 0.908 & 0.786 & 0.843 \\
FINCH  {\scriptsize(CVPR '19)}~\cite{finch}     & 0.748 & 0.677  & 0.711  \\
%\midrule
\midrule
JFAC {\scriptsize(ECCV '16)}~\cite{jfac}    & 0.690 & 0.350  & 0.460\\
TSiam {\scriptsize(FG '19)}~\cite{ssiam}                                         & {0.749} &  {0.382} & {0.506}\\
SSiam {\scriptsize(FG '19)}~\cite{ssiam}                                        & {0.766} & {0.386} & {0.514}\\
\midrule
%OSiam                               & 0.302 &  \textbf{0.652} & 0.413 \\
\textbf{CCL} (Ours with HAC)               & \textbf{0.779} &  {0.402} & \textbf{0.530} \\
\midrule
\midrule
\multicolumn{4}{c}{\textbf{\#Clusters = 40}} \\
Methods & P & R & F   \\
\midrule
FINCH  {\scriptsize(CVPR '19)}~\cite{finch} & 0.733 &  0.711 & 0.722 \\
\midrule
DIFFRAC-DeepID2$^{+}$ {\scriptsize(ICCV '11)}~\cite{jfac}  		& 0.557  & 0.213  & 0.301 \\
WBSLRR-DeepID2$^{+}$ {\scriptsize(ECCV '14)}~\cite{jfac}       	& 0.502  & 0.206  & 0.292 \\
HMRF-DeepID2$^{+}$ {\scriptsize(CVPR '13)}~\cite{jfac}        	& 0.599  & 0.230  & 0.332 \\
JFAC {\scriptsize(ECCV '16)}~\cite{jfac}                     	& 0.711  & 0.352  & 0.471 \\
TSiam {\scriptsize(FG '19)}~\cite{ssiam}                                         	    & {0.763} &  {0.362} & {0.491} \\
SSiam {\scriptsize(FG '19)}~\cite{ssiam}                                         		& {0.777} &  {0.371} & {0.502} \\
%FINCH~\cite{finch}                      & {0.733} &  \textbf{0.711} & \textbf{0.721} \\
\midrule
%OSiam                               & 0.295 & \textbf{ 0.656} &  0.407 \\
\textbf{CCL} (Ours with HAC)                             & \textbf{0.786} &  {0.392} & \textbf{0.523} \\
\bottomrule
\end{tabular}
\end{center}
\vspace{-0.6cm}
\end{table}

While previous analysis has been done at the track-level, we now report frame-level clustering performance to present a fair comparison against previous work.

\vspace{1mm}
\noindent\textbf{BBT and BF.}
We compare the results obtained with CCL to the current state-of-the-art approaches for video face clustering in Table~\ref{table:soa}.
We report clustering accuracy (\%) on two videos:
BBT-0101 and BF-0502.
Following~\cite{ssiam}, we similarly report the data source to indicate that our dataset is much harder in comparison to previous works such as~\cite{jfac,imptriplet}, that evaluate on a subset of tracks.
We can observe that CCL outperforms previous approaches, achieving $0.52\%$ and $1.33\%$ absolute improvement in  accuracy on BBT-0101 and BF-0502 respectively.

\vspace{1mm}
\noindent\textbf{ACCIO.}
% We evaluate our methods on ACCIO dataset with 36 named characters, 3243 tracks, and 166885 faces.
As common in previous works~\cite{ssiam,jfac}, we evaluate our method on the ACCIO dataset with 36 (the main cast) and 40 clusters -- see Table~\ref{table:accio}.
CCL significantly outperforms the state-of-the-art in unsupervised learning techniques and is comparable to~\cite{finch}.
Note that FINCH~\cite{finch} is not trainable.

\vspace{1mm}
\noindent\textbf{CCL vs. TSiam and SSiam.}
In Table~\ref{table:accuracy_all}, we report the frame-level clustering performance on all three datasets (as compared to track-level performance in Table~\ref{table:comparison}).
We observe that CCL outperforms TSiam, SSiam and also the base features by significant gains.
CCL operates directly on face detections (like SSiam), while TSiam requires tracks.

\vspace{2mm}
\noindent\textbf{Computational complexity.}
The time to compute FINCH partitions for BBT, BF, and ACCIO is approximately 45 seconds, and $\sim$2 minute respectively on CPU.
Training CCL for 20 epochs on BBT-0101 requires less than half an hour on a GTX 1080 GPU using the PyTorch framework.

\section{Conclusion}
\label{sec:conclusion}

We proposed a self-supervised, clustering-based contrastive learning approach for improving deep face representations.
We showed that we can train discriminative models using positive and negative pairs obtained through clustering and video level constraints that do not rely on face tracking.
% We also demonstrate that FINCH's ability to estimate the number of clusters as well as provide pure partitions plays a key role in learning a strong representation.
% Our proposed model is self-supervised that use the predicted cluster labels for training a neural network with an overall goal to improve video face clustering.
Through experiments on three challenging datasets, we showed that CCL achieves state-of-the-art performance while being computationally efficient and easily scalable.

\vspace{1mm}
\noindent
\textbf{Acknowledgments.} This work is supported by the DFG (German Research Foundation) funded PLUMCOT project.

\balance
{
%\scriptsize
%\small
\bibliographystyle{ieee}
\bibliography{main}
}

\end{document}